\documentclass{article}
\usepackage{spconf,amsmath,graphicx}
\usepackage{color}
\usepackage{subfigure}

\usepackage{multicol, multirow}

\usepackage{makecell, booktabs}
\usepackage{adjustbox}
\usepackage{float}
\usepackage{graphicx}
\usepackage{subfigure} 
\usepackage{hyperref}

\title{MORE: A Metric learning based framework for Open-domain \\ Relation Extraction}
\name{Yutong Wang$^{1}$\sthanks{Indicates equal contribution.}, Renze Lou$^{3*}$, Kai Zhang$^{2*}$, Mao Yan Chen$^{1}$, Yujiu Yang$^{1}$\sthanks{Corresponding author (\href{mailto: yang.yujiu@sz.tsinghua.edu.cn}{yang.yujiu@sz.tsinghua.edu.cn}). This work was supported in part by The National Key Research and Development Program of China (No. 2020YFB1708200) and the Guangdong Basic and Applied Basic Research Foundation (No. 2019A1515011387).}}
\address{$^{1}$Tsinghua Shenzhen International Graduate School, Tsinghua University \\ $^{2}$Department of Computer Science and Technology, Tsinghua University, Beijing, China\\
$^{3}$Department of Computer Science, Zhejiang University City College, Hangzhou, China\\}
\begin{document}
%
\maketitle
\begin{abstract}
Open relation extraction (OpenRE) is the task of extracting relation schemes from open-domain corpora. Most existing OpenRE methods either do not fully benefit from high-quality labeled corpora or can not learn semantic representation directly, affecting downstream clustering efficiency. To address these problems, in this work, we propose a novel learning framework named MORE (\textbf{M}etric learning-based \textbf{O}pen \textbf{R}elation \textbf{E}xtraction). The framework utilizes deep metric learning to obtain rich supervision signals from labeled data and drive the neural model to learn semantic relational representation directly. Experiments result in two real-world datasets show that our method outperforms other state-of-the-art baselines. Our source code is available on Github\footnote{\url{https://github.com/RenzeLou/MORE}}.
\end{abstract}


\section{Introduction}
\label{sec:intro}
Relation extraction (RE) is an important NLP task that aims to detect and categorize semantic relations between entities and has many applications, such as knowledge graph construction \cite{suchanek2007yago}, information retrieval \cite{xiong2017explicit}, and logic reasoning \cite{socher2013reasoning}. However, with the rapid emergence of novel knowledge, the corresponding relationship types in open-domain corpora are also increasing, which is challenging for RE to handle. Thus, OpenRE is proposed to solve this problem  \cite{banko2007open}, which aims at extracting relational schemes from the open-domain corpus without predefined relation types. 

Existing OpenRE methods are divided into two main categories: tagging-based and clustering-based. The tagging-based methods formulate OpenRE as a sequence labeling problem \cite{banko2007open,banko2008tradeoffs}, but these methods often extract surface forms and are difficult to be utilized for downstream tasks. Meanwhile, many efforts are devoted to exploring clustering-based methods that cluster semantic patterns into certain relation types, such as \cite{yao2012unsupervised, elsahar2017unsupervised}, yet those schemes are laborious and time-consuming due to the high dependence on rich features. Recently, neural networks began to be exploited in clustering-based OpenRE tasks to alleviate the above issues. For example, Hu et al.~\cite{hu2020selfore} utilize a neural model to capture self-supervision signals and detect novel instances in an open scene. Gao et al.~\cite{gao2020neural} propose a siamese network, which accumulates novel types with its few-shot instances. Besides these bootstrapping methods, another supervised scheme learns the similarity metrics from labeled instances and further transfer the relational knowledge to open-domain corpora, namely Relational Siamese Networks (RSNs) \cite{wu2019open}. However, RSNs target at learning a similarity classifier rather than building relational representations directly. Thus this may affect the speed and efficiency of downstream clustering.

To address this issue, we propose a new supervised learning scheme, which applies deep metric learning to clustering-based OpenRE and build semantic embeddings directly. From our insight view, the essential objective of the cluster-based OpenRE algorithm is to distinguish relational texts and detect the novel classes. Thus, learning semantic representations from sentences is crucial for downstream clustering. Furthermore, metric learning algorithms aim to learn the representations and acquire semantic space directly. Therefore, it is rational to take metric learning into count. However, most prevailing deep metric learning methods, such as triplet loss \cite{hoffer2015deep}, N-pair-mc \cite{sohn2016improved}, or Proxy-NCA \cite{movshovitz2017no}, always suffer from poor supervision signals from the limited number of data points, which may harm the performance of novel relation detection. Inspired by \cite{wang2019ranked}, we take Ranked List Loss (RLL) as our choice, which can capture set-based supervision signals and gain richer information. In addition, we design virtual adversarial training to enhance the model's robustness and deal with the noise in open scenes. Experiments demonstrate that MORE can learn more semantic representations and achieve state-of-the-art results on real-world datasets. To sum up, the main contributions of us are as follows:

1.~To the best of our knowledge, we are the first to adopt deep metric learning in OpenRE tasks and propose a new learning framework MORE that can learn semantic representations and be used for downstream clustering directly. Meanwhile, we also design virtual adversarial training on our model to smooth the semantic space.  

2.~Experiments illustrate that the proposed MORE achieves state-of-the-art performance on real-world datasets, even if the imbalance distribution presents in the test set. Moreover, the visual analysis also demonstrates its excellent ability of representation learning.

\section{METHODOLOGY}
\begin{figure}[t]
	\begin{center}
		\centering
		\includegraphics[width=1.0\linewidth]{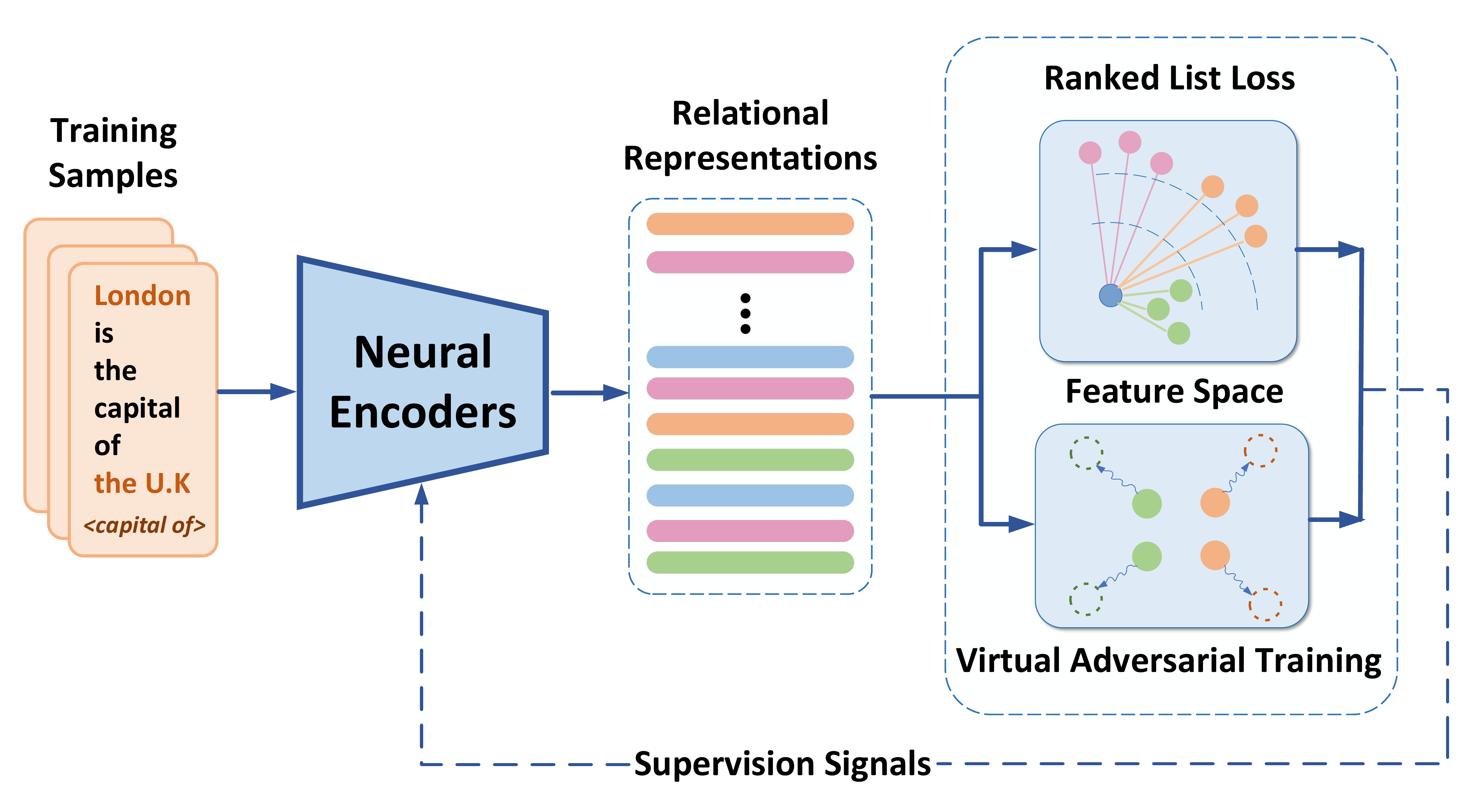}
	\end{center}
	\vspace{-2em}
	\caption{Overall architecture of MORE.}
	\label{fig:architectures}
	\vspace{-0.3cm}
\end{figure}

As shown in Figure \ref{fig:architectures}, the proposed framework MORE exploits the neural encoders to extract relational representations and use them to calculate the Ranked List Loss (RLL). Besides, we set virtual adversarial training to smooth the semantic space.
\subsection{Neural Encoders}
As a vital component of our method, neural encoder aim at learning semantic representations of relation types. In this paper, we have experimented with two different encoders.


\textbf{CNN} \quad Following \cite{wu2019open}, we take the CNN encoder as our first choice, which includes an embedding layer followed by a one-dimensional convolutional layer and a max-pooling layer. The model setting we used is the same as \cite{wu2019open}.

\textbf{BERT} \quad Inspired by SelfORE \cite{hu2020selfore}, which exploits the pre-trained language model, we also choose BERT \cite{devlin2018bert} as our encoder and follow the operation proposed by \cite{soares2019matching} to fit our OpenRE task better. Specifically, for a sentence $\mathcal{S} = \{s_1,..,s_T\}$, where $s$ indicates the token and $T$ is the length of $\mathcal{S}$. We insert four special tokens before and after each entity mentioned in a sentence and get a new sequence:
\begin{align}
\mathcal{S} = [s_1,..,[E1_{start}],s_p,..,s_q,[E1_{end}], \notag\\
..,[E2_{start}],s_k,..,s_l,[E2_{end}],..,s_T] 
\end{align}
We use this sequence as the input of BERT, and we concatenate the last hidden state of BERT's outputs corresponding to $[E1_{start}], [E2_{start}]$, take it as our relational representation.

After obtaining the relational representations, we then use a fully-connected layer to map representations into hidden states. Next, we apply $L_{2}$ normalization on every hidden state and use Euclidean distance to measure the similarity among them.

\subsection{Ranked List Loss}
We utilize deep metric learning to optimize the semantic space after defining the distance metric between representations in normalized Euclidean space. Unlike most other metric learning methods, such as triplet loss \cite{hoffer2015deep}, N-pair-mc \cite{sohn2016improved}, which are limited from point-based or pair-based information. Ranked List Loss (RLL) explores the set-based similarity structure from the training batch, and obtain richer supervision signals. For an anchor selected from the training batch, RLL rank the similarity of all the same type (positive) points before the different categories (negative) points and preserve a margin between them.

\begin{figure}[tp]
     \centering
     \includegraphics[width=0.9\columnwidth]{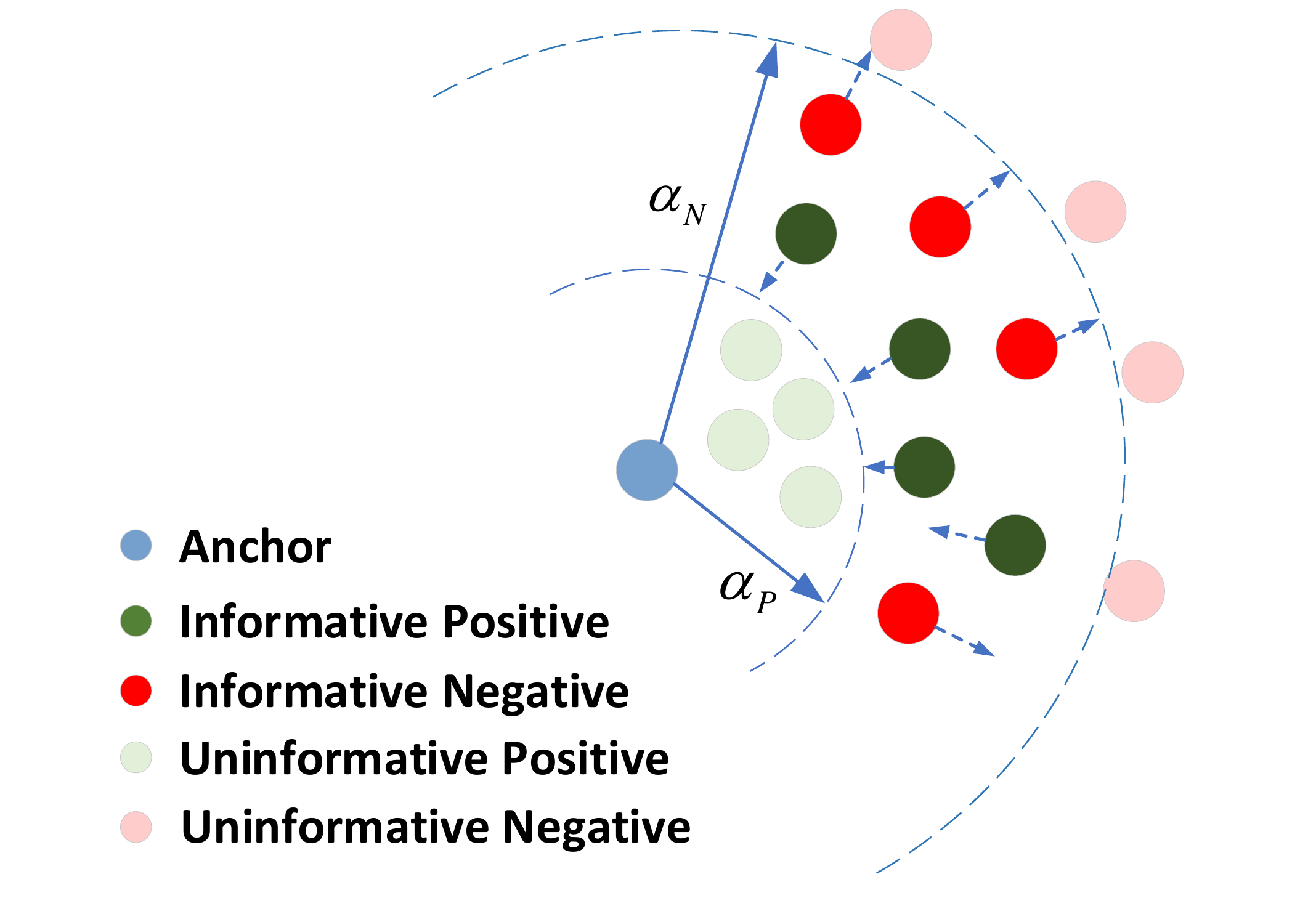}
     \caption{Illustration of the ranked list loss.}
     \label{fig:rll}
     \vspace{-0.3cm} 
\end{figure}

To be specific, given a set of normalized relation representations $\mathcal{R} = \{r_1,..,r_n\} $ , where $n$ indicates the total number of labeled sentences. We sample $m (m\leq n)$ instances of $c$ types randomly from $\mathcal{R}$ and group them into a batch $\mathcal{B} = \{r_k,..,r_{k+l}\}$. Intuitively, given an instance(anchor) $r_i$ in $\mathcal{B}$, we expect the positive points in $\mathcal{B}$ can be gathered together while those negative points are the opposite. Then, we calculate the following formula:
\begin{align}
    \mathcal{L}(r_i,\mathcal{B};f)=\sum_{r_j\in \mathcal{B},j\neq i} [(1-y_{ij}) [\alpha_{N}-d_{ij}]_{+} \notag\\
    +y_{ij}[d_{ij}-\alpha_{P}]_{+}]
\end{align}
where $f$ is model parameters, $y$ indicates the relation type, $y_{i j}=1$ if $r_j$ is a positive point, and $y_{i j}=0$ otherwise. $d_{ij}$ denotes the Euclidean distance between two points. $\alpha_{P},\alpha_{N}$ represent the positive and negative boundary respectively, $[.]_{+}$ denote the hinge function. As shown in Figure \ref{fig:rll}, those positive instances outside the $\alpha_{P}$ will be pulled closer, while those negative points within the $\alpha_{N}$ will be pushed farther. The remaining uninformative points that already meet our objective will not be taken into count because of the hinge function.

More concisely, for $r_i$ in $\mathcal{B}$, let's define $\mathcal{L}_{P}(r_i,\mathcal{B};f)$ as the total loss of all informative positive points, and $\mathcal{L}_{N}(r_i,\mathcal{B};f)$ is the sum of all informative negative samples loss, the optimization objective function can be summarized as below:
\begin{small}
\begin{align}
    \mathcal{L}_{RLL}\left(\mathcal{B};f\right)=\sum_{r_i\in \mathcal{B}}\left[\left(1-\lambda\right)\mathcal{L}_{P}(r_i,\mathcal{B};f)+\lambda\mathcal{L}_{N}(r_i,\mathcal{B};f)\right] \label{eq:overall}
\end{align}
\end{small}
Here, the $\lambda$ is used to control the balance between $\mathcal{L}_{P}(r_i,\mathcal{B};f)$ and $\mathcal{L}_{N}(r_i,\mathcal{B};f)$, which we set to 0.5 practically.

Usually, given $r_i$ as a anchor, numerous informative negative points can be found in $\mathcal{B}$. To deal with the magnitude difference laying in the negative loss, we follow \cite{wang2019ranked}, weighting the negative examples according to the values of their loss:
\begin{align}
    w_{ij}=\exp [T_n \ast (\alpha-d_{ij})], r_j\in \mathcal{R}_{i}^{\mathcal{N}}
\end{align}
Where $\mathcal{R}_{i}^{\mathcal{N}}$ represents the set of informative negative samples of $r_i$, and $T_n$ indicates the temperature which controls the degree of weighting these samples. For example, when $T_n$ is 0, every instance will be treated equally. And setting $T_n$ to $+\infty$ when devoting almost all attention to the hardest sample.  

Consequently, the $\mathcal{L}_{N}(r_i,\mathcal{B};f)$ in (\ref{eq:overall}) can be updated as:
\begin{align}
    \mathcal{L}_{N}(r_i,\mathcal{B};f)=\sum_{r_j\in \mathcal{R}_{i}^{\mathcal{N}}}[\frac{w_{ij}}{\sum_{r_j\in 
    \mathcal{R}_{i}^{\mathcal{N}}}w_{ij}}(\alpha_N-d_{ij})]
\end{align}
After the model retrieve the supervision signals of $\mathcal{B}$, we sample the next $m$ instances from $\mathcal{R}$ and build a new group iteratively.
\subsection{Virtual adversarial training}
To alleviate the diversity and noise present in the text of open scenarios, we set virtual adversarial training (VAT) \cite{miyato2018virtual} to smooth the semantic space and enhance the model's robustness.

Specifically, for a given sentence $S$ and its original representation $r$, we first generate a normalized perturbation on the word embeddings within $S$ randomly and take this disturbed embeddings as the input of encoder to build a new representation $\tilde{r}$. Next, we calculate the Euclidean distance between $r$ and $\tilde{r}$ and take the gradient $g$ of the distance. Then, we regard $\epsilon$ times normalized $g$ as the worst-case perturbation $\xi$, where $\epsilon$ is a small decimal we set to 0.02 in all our experiments. Finally, we use $\xi$ to disturb the representation build by the model and penalize our model with the VAT loss:
\begin{align}
    \mathcal{L}_{adv}(\mathcal{B};f)=\frac{1}{m}\sum_{i=1}^{m} D( F(S_i;f), F(S_i+\xi_i;f) )
\end{align}
$F(S_i+\xi_i;f)$ denotes the distributed representation encoded by the neural model, while $F(S_i;f)$ is the original one and $D$ calculates the Euclidean distance between two representations. Intuitively, we expect any representation $r_i$ in $\mathcal{B}$ is stable as possible under such worst-case perturbations.

Thence, the final objective loss function can be written as:
\begin{align}
    \mathcal{L}(\mathcal{B};f)= \mathcal{L}_{RLL}(\mathcal{B};f)+\beta \mathcal{L}_{adv}(\mathcal{B};f)
\end{align}
Where $\beta$ is a factor that indicates the weight of virtual adversarial training, and we set it to 1, same as \cite{miyato2018virtual}.

\section{experiment}
In this section, we conduct some experiments on real-world RE datasets to show the effectiveness of our model.
\subsection{Experiment Setting}
\textbf{Datasets} We perform our experiments on two datasets. The first one is FewRel \cite{han2018fewrel}, derived from Wikipedia and annotated by crowd workers. Different from most other datasets, the entity pair of each instance in FewRel is unique, which makes the model unable to obtain shortcuts by memorizing the entities. Following \cite{wu2019open}, we choose 64 relations as the train set and randomly select 16 relations with 1600 instances as the test set; the remaining sentences are validation set. The other one is NYT+FB \cite{sandhaus2008new}\cite{bollacker2008freebase}, which is built by distant supervision. To fit the supervision setting, we divide the original dataset again and generate NYT+FB-sup. Usually, the relations which occur frequently are common categories, and those relations with rare instances are insufficient to be regarded as novel types. Therefore, we select relations with the number of instances between 20 and 2000 as novel relations. We finally obtain 72 novel relations equally divided between the test and validation set, leaving 190 relations as the train set, which contain both common and extremely rare types to simulate an unbalanced real environment.

\noindent\textbf{Evaluation Metrics} We adopt $\mathrm{B}^{3}$ $ F_1$ score  \cite{bagga1998algorithms} as our metrics, which is widely used in previous works \cite{wu2019open,hu2020selfore,elsahar2017unsupervised,marcheggiani2016discrete}. $F_1$ calculates the harmonic mean of precision and recall, its value is more affected by the lower one, which can fairly demonstrate the performance of the model.

\noindent\textbf{Implementation Details} 
In all our experiments, we choose Adam \cite{kingma2014adam} for our optimization. We fix the learning rate with 0.003 and 0.00001 on CNN and BERT respectively. For the content of a batch $\mathcal{B}$, we set the number of relation types $c$ to 24, each type with 10 instances. For the hyperparameter $\alpha_P,\alpha_N$, we set 0.8 and 1.2 separately, thus preserving a margin of 0.4 between these two boundaries. The temperature factor $T_n$ in this work is 10. On FewRel, we choose K-Means \cite{hartigan1979algorithm} as our downstream clustering algorithm and set the number of clusters with 16. And on NYT+FB-sup, to deal with the imbalance, we choose Mean-Shift \cite{cheng1995mean} instead of K-means, which can automatically find clusters based on spatial density.

\subsection{Result}
We compare with four baselines \cite{wu2019open,hu2020selfore,marcheggiani2016discrete,elsahar2017unsupervised} on two datasets. All these models are evaluated on the test set to show their performance.

\begin{table}[!t]
    \centering
    \tiny
    \adjustbox{max width=\linewidth,width=0.95\linewidth,center=\linewidth}{
    \begin{tabular}{l|ccc}
        \Xhline{0.5pt}
         \multirow{3}{*}{\textbf{Method}} & \multicolumn{3}{c}{\textbf{FewRel}}\\
          & Prec. & Rec. & F1\\    
        \Xhline{0.3pt}
         VAE \cite{marcheggiani2016discrete}  & 17.9 & 69.7 & 28.5\\
         RW-HAC \cite{elsahar2017unsupervised} & 31.8 & 46.0 & 37.6\\
         SelfORE \cite{hu2020selfore} & 50.8 & 51.6 & 51.2  \\
         RSNs \cite{wu2019open} & 48.9 & 77.5 & 59.9\\
        \hline \hline
        MORE(CNN) & 56.6 & 60.3 & 58.4\\
        MORE(CNN)+VAT & 57.1 & 68.0 & 62.0\\
        MORE(BERT) & \textbf{70.1} & \textbf{79.6} & \textbf{74.5}\\
        
        \Xhline{0.5pt}
    \end{tabular}}
    \caption{The results on FewRel.}
    \label{tab:FewRel}
\end{table}

\begin{table}[!t]
    \centering
    \tiny
    \adjustbox{max width=\linewidth,width=0.95\linewidth,center=\linewidth}{
    \begin{tabular}{l|ccc}
         \Xhline{0.5pt}
         \multirow{3}{*}{\textbf{Method}} & \multicolumn{3}{c}{\textbf{NYT+FB-sup}}\\
          & Prec. & Rec. & F1\\    
        \Xhline{0.3pt} 
         VAE \cite{marcheggiani2016discrete} & 20.3 & 40.7 & 27.1 \\
         RW-HAC \cite{elsahar2017unsupervised} & 25.2 & 33.9 & 28.9\\
         SelfORE \cite{hu2020selfore} & 30.9 & 46.4 & 37.1\\
         RSNs \cite{wu2019open} & 31.1 & \textbf{52.0} & 38.8\\
        \hline
        \hline
        MORE(CNN) & 36.4 & 48.9 & 41.8\\
        MORE(CNN)+VAT & 39.1 & 49.1 & 43.5\\
        MORE(BERT) & \textbf{48.7} & 50.8 & \textbf{49.7}\\
        \Xhline{0.5pt}
    \end{tabular}}
    \caption{The results on NYT+FB-sup.}
    \label{tab:NYT}
    \vspace{-0.25cm} 

\end{table}

The main results can be seen in Table \ref{tab:FewRel} and Table \ref{tab:NYT}, the scores of all algorithm are the highest one among the statistical testings(some borrowed from the original paper). We can draw the following conclusions:

(1) Benefit by the rich supervision signals come from the labeled data, MORE outperforms all unsupervised or self-supervised methods on both datasets, such as SelfORE \cite{hu2020selfore} which used to achieve admirable results on NYT+FB. The results indicate the effectiveness of prior knowledge transfer, which is conducive to novel type-detection in open scenarios.

(2) MORE(BERT) outperforms RSNs on precision and $F_1$. However, the superiority is not obvious on recall because the clustering method adopted by RSNs is Louvain \cite{blondel2008fast}, which constantly produces coarse-grained clustering results, as mentioned in \cite{wu2019open}. It is also worth noting that, with only a CNN that is the same as RSNs used, MORE(CNN) can also achieve notable results, demonstrating the capability of our metric learning-based scheme.

(3) With the pre-trained language model, MORE's performance has been significantly improved on both two datasets. The result reflects that benefit from the plentiful information captured by RLL, BERT can be fine-tuned to pay more attention to the contextualized relation extraction, thereby constructing more richly semantic embeddings.

(4) The $F_1$ scores of all methods on NYT+FB-sup are lower than the results on FewRel, which is mainly because the noisy training signals from distance supervision and the data is unevenly distributed. However, even in a tough setting, MORE can still maintain a better performance than others. The result proves that MORE is robust to noise in the dataset and able to distinguish those ambiguous novels and rare classes in open-domain corpora.

(5) Virtual adversarial training has not achieved a significant improvement compared with RSNs, indicating the smoothing difference between Euclidean space and probability distribution space. We will leave this optimization problem for our future work.

Besides, due to the massive consumption of computing resources for fine-tuning of BERT, we do not apply virtual adversarial training on MORE(BERT), which we will resolve in the future.

\subsection{Visual Analysis}

\begin{figure}[htbp]
\centering
\vspace{-0.5cm} 
\subfigure[MORE]{
\includegraphics[width=4.0cm]{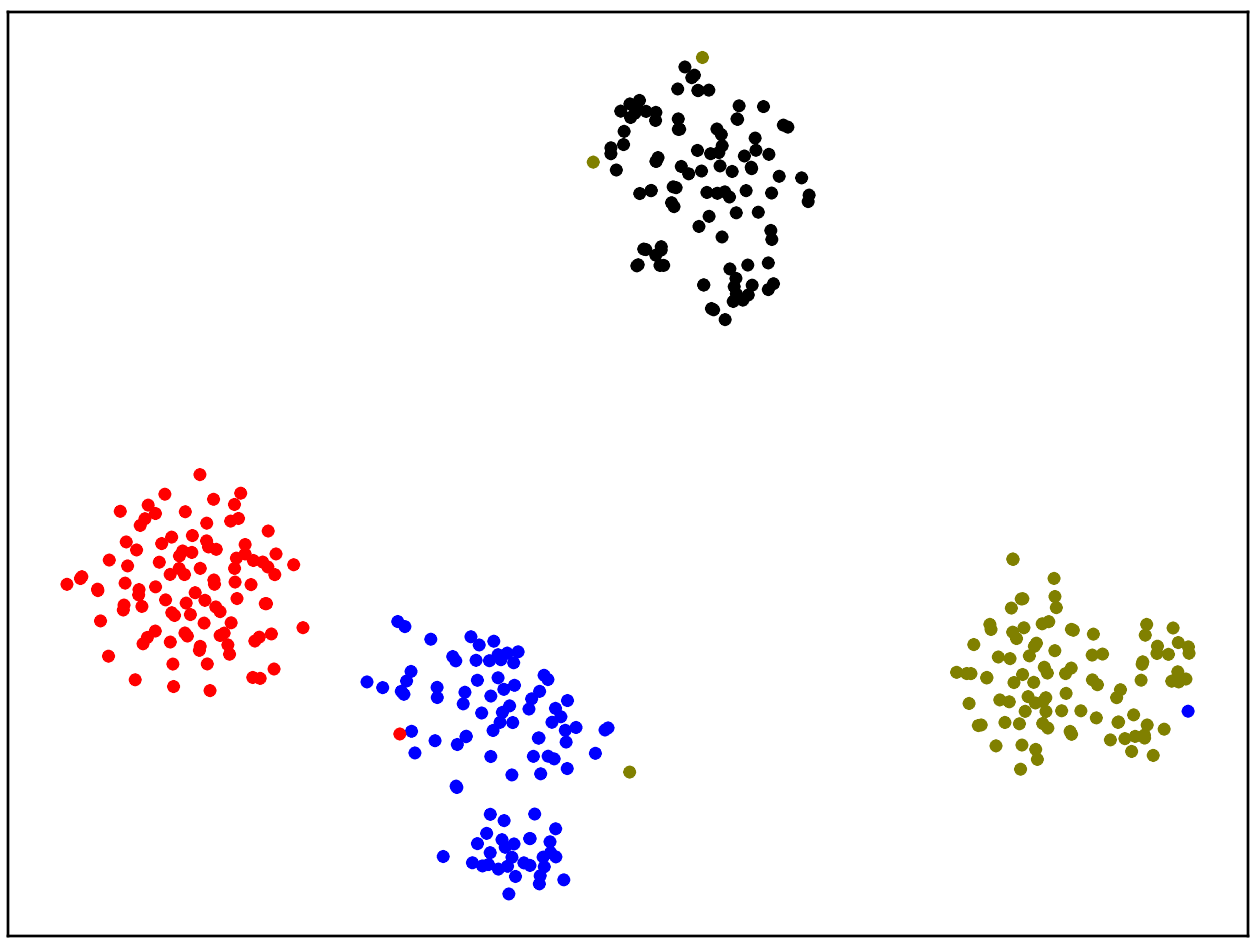}
}
\subfigure[RSNs]{
\includegraphics[width=4.0cm]{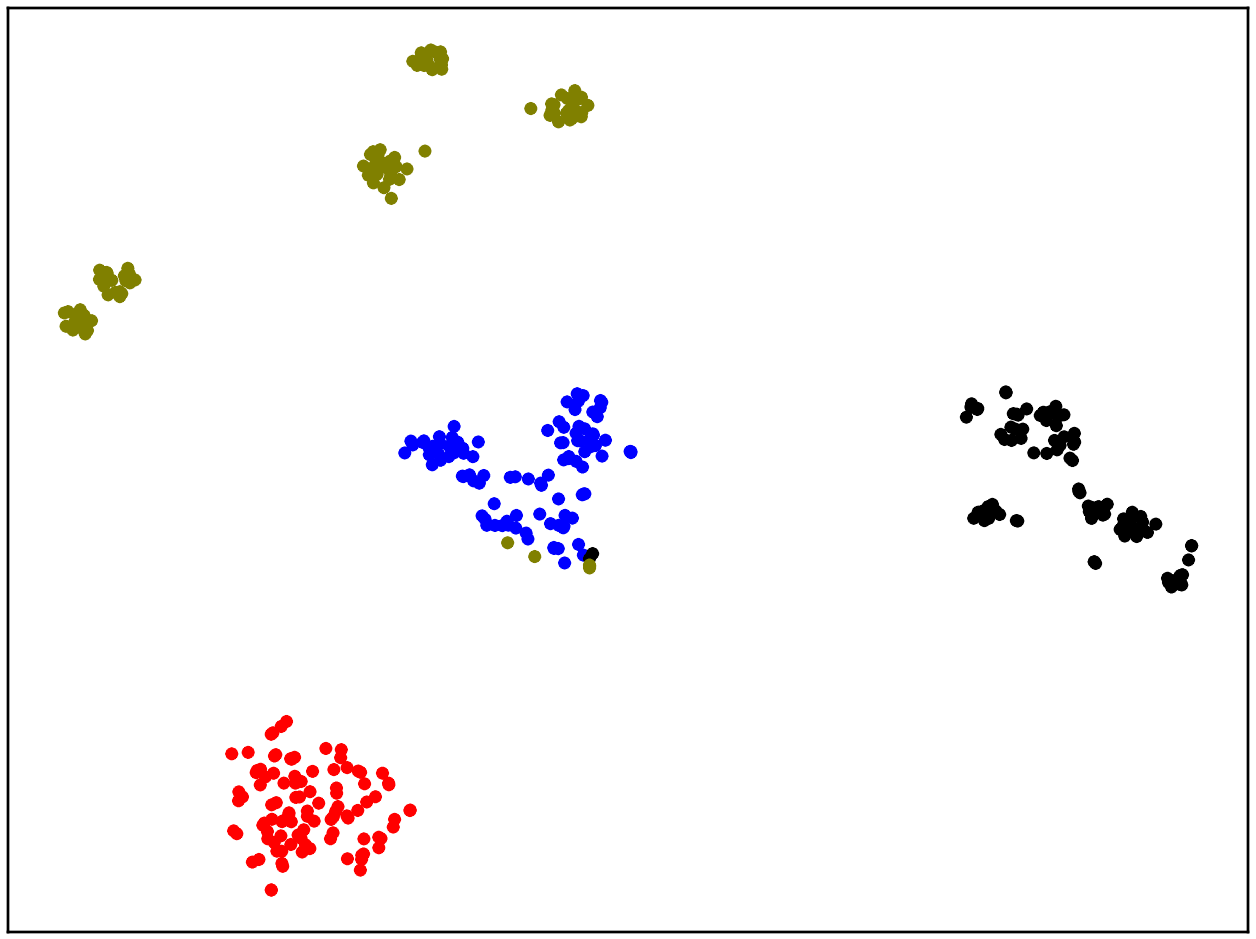}
}
\vspace{-0.35cm}
\caption{The t-SNE visualization on FewRel.}
\label{fig:t-SNE1.0}
\end{figure}
\vspace{-0.23cm} 

In order to intuitively demonstrate the capability of MORE on semantic representation learning, we visualize the semantic space of relational representations with t-SNE \cite{maaten2008visualizing}. Specifically, We randomly sample 4 relation types from the test set of FewRel, 100 instances per type, and construct representations from these instances on MORE(CNN) and RSNs respectively, colored according to their ground-truth types. 

As shown in Figure \ref{fig:t-SNE1.0}, the semantic space of MORE is more distinguishable that almost all four types can preserve the intraclass similarity within a hypersphere while leaving a distinct margin between any two categories. In contrast, RSNs attempt to shrink the representations of the same type into one point, thus the distribution of points in each cluster is denser than MORE. However, excessive attention to the similarity between point pairs may drop the intraclass similarity structure, so it is easier for RSNs to divide samples of the same category into multiple subcategories.






\section{Conclusion}
In this paper, we propose a novel framework for open-domain relation extraction, which utilizes deep metric learning to drive the neural model to learn relational representations directly, thereby conducive to the efficiency of downstream clustering. Moreover, we set virtual adversarial training to enhance the robustness of the neural encoder. Our experiments demonstrate the capability of our scheme on relational representation learning and novel relation detection.
\label{sec:conclusion}

\clearpage
\small
\bibliographystyle{newbib}
\bibliography{refs}

\begin{thebibliography}{10}

\bibitem{suchanek2007yago}
Fabian~M Suchanek, Gjergji Kasneci, and Gerhard Weikum,
\newblock ``Yago: a core of semantic knowledge,''
\newblock in {\em Proceedings of the 16th international conference on World
  Wide Web}, 2007, pp. 697--706.

\bibitem{xiong2017explicit}
Chenyan Xiong, Russell Power, and Jamie Callan,
\newblock ``Explicit semantic ranking for academic search via knowledge graph
  embedding,''
\newblock in {\em Proceedings of the 26th international conference on world
  wide web}, 2017, pp. 1271--1279.

\bibitem{socher2013reasoning}
Richard Socher, Danqi Chen, Christopher~D Manning, and Andrew Ng,
\newblock ``Reasoning with neural tensor networks for knowledge base
  completion,''
\newblock in {\em Advances in neural information processing systems}, 2013, pp.
  926--934.

\bibitem{banko2007open}
M~Banko, MJ~Cafarella, S~Soderland, M~Broadhead, and O~Etzioni,
\newblock ``Open information extraction from the web in: Proceedings of the
  20th international joint conference on artificial intelligence,'' 2007.

\bibitem{banko2008tradeoffs}
Michele Banko and Oren Etzioni,
\newblock ``The tradeoffs between open and traditional relation extraction,''
\newblock in {\em Proceedings of ACL-08: HLT}, 2008, pp. 28--36.

\bibitem{yao2012unsupervised}
Limin Yao, Sebastian Riedel, and Andrew McCallum,
\newblock ``Unsupervised relation discovery with sense disambiguation,''
\newblock in {\em Proceedings of the 50th Annual Meeting of the Association for
  Computational Linguistics (Volume 1: Long Papers)}, 2012, pp. 712--720.

\bibitem{elsahar2017unsupervised}
Hady Elsahar, Elena Demidova, Simon Gottschalk, Christophe Gravier, and
  Frederique Laforest,
\newblock ``Unsupervised open relation extraction,''
\newblock in {\em European Semantic Web Conference}. Springer, 2017, pp.
  12--16.

\bibitem{hu2020selfore}
Xuming Hu, Lijie Wen, Yusong Xu, Chenwei Zhang, and Philip~S Yu,
\newblock ``Selfore: Self-supervised relational feature learning for open
  relation extraction,''
\newblock {\em arXiv preprint arXiv:2004.02438}, 2020.

\bibitem{gao2020neural}
Tianyu Gao, Xu~Han, Ruobing Xie, Zhiyuan Liu, Fen Lin, Leyu Lin, and Maosong
  Sun,
\newblock ``Neural snowball for few-shot relation learning.,''
\newblock in {\em AAAI}, 2020, pp. 7772--7779.

\bibitem{wu2019open}
Ruidong Wu, Yuan Yao, Xu~Han, Ruobing Xie, Zhiyuan Liu, Fen Lin, Leyu Lin, and
  Maosong Sun,
\newblock ``Open relation extraction: Relational knowledge transfer from
  supervised data to unsupervised data,''
\newblock in {\em Proceedings of the 2019 Conference on Empirical Methods in
  Natural Language Processing and the 9th International Joint Conference on
  Natural Language Processing (EMNLP-IJCNLP)}, 2019, pp. 219--228.

\bibitem{hoffer2015deep}
Elad Hoffer and Nir Ailon,
\newblock ``Deep metric learning using triplet network,''
\newblock in {\em International Workshop on Similarity-Based Pattern
  Recognition}. Springer, 2015, pp. 84--92.

\bibitem{sohn2016improved}
Kihyuk Sohn,
\newblock ``Improved deep metric learning with multi-class n-pair loss
  objective,''
\newblock in {\em Advances in neural information processing systems}, 2016, pp.
  1857--1865.

\bibitem{movshovitz2017no}
Yair Movshovitz-Attias, Alexander Toshev, Thomas~K Leung, Sergey Ioffe, and
  Saurabh Singh,
\newblock ``No fuss distance metric learning using proxies,''
\newblock in {\em Proceedings of the IEEE International Conference on Computer
  Vision}, 2017, pp. 360--368.

\bibitem{wang2019ranked}
Xinshao Wang, Yang Hua, Elyor Kodirov, Guosheng Hu, Romain Garnier, and Neil~M
  Robertson,
\newblock ``Ranked list loss for deep metric learning,''
\newblock in {\em Proceedings of the IEEE Conference on Computer Vision and
  Pattern Recognition}, 2019, pp. 5207--5216.

\bibitem{devlin2018bert}
Jacob Devlin, Ming-Wei Chang, Kenton Lee, and Kristina Toutanova,
\newblock ``Bert: Pre-training of deep bidirectional transformers for language
  understanding,''
\newblock {\em arXiv preprint arXiv:1810.04805}, 2018.

\bibitem{soares2019matching}
Livio~Baldini Soares, Nicholas FitzGerald, Jeffrey Ling, and Tom Kwiatkowski,
\newblock ``Matching the blanks: Distributional similarity for relation
  learning,''
\newblock {\em arXiv preprint arXiv:1906.03158}, 2019.

\bibitem{miyato2018virtual}
Takeru Miyato, Shin-ichi Maeda, Masanori Koyama, and Shin Ishii,
\newblock ``Virtual adversarial training: a regularization method for
  supervised and semi-supervised learning,''
\newblock {\em IEEE transactions on pattern analysis and machine intelligence},
  vol. 41, no. 8, pp. 1979--1993, 2018.

\bibitem{han2018fewrel}
Xu~Han, Hao Zhu, Pengfei Yu, Ziyun Wang, Yuan Yao, Zhiyuan Liu, and Maosong
  Sun,
\newblock ``Fewrel: A large-scale supervised few-shot relation classification
  dataset with state-of-the-art evaluation,''
\newblock {\em arXiv preprint arXiv:1810.10147}, 2018.

\bibitem{sandhaus2008new}
Evan Sandhaus,
\newblock ``The new york times annotated corpus,''
\newblock {\em Linguistic Data Consortium, Philadelphia}, vol. 6, no. 12, pp.
  e26752, 2008.

\bibitem{bollacker2008freebase}
Kurt Bollacker, Colin Evans, Praveen Paritosh, Tim Sturge, and Jamie Taylor,
\newblock ``Freebase: a collaboratively created graph database for structuring
  human knowledge,''
\newblock in {\em Proceedings of the 2008 ACM SIGMOD international conference
  on Management of data}, 2008, pp. 1247--1250.

\bibitem{bagga1998algorithms}
Amit Bagga and Breck Baldwin,
\newblock ``Algorithms for scoring coreference chains,''
\newblock in {\em The first international conference on language resources and
  evaluation workshop on linguistics coreference}. Citeseer, 1998, vol.~1, pp.
  563--566.

\bibitem{marcheggiani2016discrete}
Diego Marcheggiani and Ivan Titov,
\newblock ``Discrete-state variational autoencoders for joint discovery and
  factorization of relations,''
\newblock {\em Transactions of the Association for Computational Linguistics},
  vol. 4, pp. 231--244, 2016.

\bibitem{kingma2014adam}
Diederik~P Kingma and Jimmy Ba,
\newblock ``Adam: A method for stochastic optimization,''
\newblock {\em arXiv preprint arXiv:1412.6980}, 2014.

\bibitem{hartigan1979algorithm}
John~A Hartigan and Manchek~A Wong,
\newblock ``Algorithm as 136: A k-means clustering algorithm,''
\newblock {\em Journal of the royal statistical society. series c (applied
  statistics)}, vol. 28, no. 1, pp. 100--108, 1979.

\bibitem{cheng1995mean}
Yizong Cheng,
\newblock ``Mean shift, mode seeking, and clustering,''
\newblock {\em IEEE transactions on pattern analysis and machine intelligence},
  vol. 17, no. 8, pp. 790--799, 1995.

\bibitem{blondel2008fast}
Vincent~D Blondel, Jean-Loup Guillaume, Renaud Lambiotte, and Etienne Lefebvre,
\newblock ``Fast unfolding of communities in large networks,''
\newblock {\em Journal of statistical mechanics: theory and experiment}, vol.
  2008, no. 10, pp. P10008, 2008.

\bibitem{maaten2008visualizing}
Laurens van~der Maaten and Geoffrey Hinton,
\newblock ``Visualizing data using t-sne,''
\newblock {\em Journal of machine learning research}, vol. 9, no. Nov, pp.
  2579--2605, 2008.

\end{thebibliography}
\end{document}